# FlameGS: Reconstruct flame light field via Gaussian Splatting

Yunhao Shui[a], Fuhao Zhang[a], Can Gao[a], Hao Xue[a], Zhiyin Ma[a], Gang Xun[a], Xuesong Li*[a]

[a]School of Mechanical Engineering, Shanghai Jiao Tong University, 800 Dongchuan Road, Shanghai, 200240, PR China

## ABSTRACT

To address the time-consuming and computationally intensive issues of traditional ART algorithms for flame combustion diagnosis, inspired by flame simulation technology, we propose a novel representation method for flames. By modeling the luminous process of flames and utilizing 2D projection images for supervision, our experimental validation shows that this model achieves an average structural similarity index of 0.96 between actual images and predicted 2D projections, along with a Peak Signal-to-Noise Ratio of 39.05. Additionally, it saves approximately 34 times the computation time and about 10 times the memory compared to traditional algorithms.

## 1. INTRODUCTION

Flame combustion diagnostics are essential in engines and industrial applications. Traditional methods like Computed Tomography of Chemiluminescence (CTC) reconstruct 3D flame fields from 2D projections.

However, commonly used methods like Algebraic Reconstruction Technique (ART)[1] relying on voxel-based representation are low efficiency and lack of accuracy. Specifically, ART suffers from high computational and memory demand, as voxel complexity grows cubically with resolution. And constructing the weight matrix between 2D pixels and 3D voxels is both resource-intensive and time-consuming. Furthermore, the predefined voxel boundaries strictly constrain the reconstruction area, leaving material outside the voxel unaccounted for and causing artifacts.

Deep learning approaches have also explored voxel-based representations, such as CNN-based models[2], which reconstruct 3D flame volumes from 2D projection images. Despite their advantages, these methods remain limited by voxel constraints and require large datasets with 3D voxel ground truth for supervision, which is hard to acquire.

While NeRF[3], a differentiable rendering technique based on ray tracing has excluded the dependence on voxel by implicitly represents 3D volumetric structures as MLP (Multilayer Perceptron), it faces several challenges. Implicit representation cannot be visualized directly, and it has tons of parameters to be optimized, which is also computational-consuming, with training time ranges from several hours to several days.

Existing approaches are often constrained by voxel-based representations or other low-efficiency methods, prompting the need for a more efficient, scalable, and interpretable solution. Drawing inspiration from Gaussian functions—widely applied in flame modeling and simulation[8,9], we propose a novel representation of the flame volume as a set of 3D Gaussians [5], entirely free of voxels. This representation enables fast, memory-efficient rendering and facilitates a more interpretable 3D structure. By optimizing the parameters of these Gaussians directly from 2D projections, our method avoids the need for 3D ground truth data and significantly reduces the computational complexity. We validated our method using experimental data, achieving a better reconstruction accuracy, and the training times nearly 30 times faster and training memory usage 10 times lower than the ART algorithm.

In summary, our contributions can be summarized as follows:

  1. We proposed an efficient and high accuracy flame volumetric representation method based on 3D Gaussians.

  2. We proposed a ray tracing-based random initialization method, which accelerates model convergence and improves reconstruction quality.

  3. We validated our model through experiments, demonstrating the strong performance of our approach and the effectiveness of the proposed components.

*xuesonl@sjtu.edu.cn

## 2. RELATED WORKS

Various improvements to the ART have been proposed to mitigate its computational challenges. The Simultaneous Algebraic Reconstruction Technique (SART) [6] optimizes voxels simultaneously, while the Multiplicative ART (MART) [7] improves numerical stability by replacing additive iteration with multiplicative iteration. However, due to its voxel representation, all these ART-based methods rely on a massive weight matrix $W$, calculated via ray tracing to link each voxel to each pixel in the projected images. Detailly, the shape of $W$ is typically $(R_x \times R_y \times R_z, N \times H \times W)$, where $R_x$, $R_y$, and $R_z$ are voxel resolutions, and $N$, $H$, and $W$ represent the number. The shape of the images. $W$ is often of trillion-parameter scale, requiring tens of GBs and significant computational time for optimization. The reconstructed area is restricted to the predefined voxel boundaries, limiting its ability to capture regions outside this predefined space.

To address the memory demands of the matrix, H. Pan et al.[8] introduced WERNET, which replaces the matrix with a parameter-learnable neural network, significantly reducing parameter count. And Huang et al.[2] proposed using a CNN to train a generalized model that reconstructs 3D flame volumes directly from surrounding projection images. However, this approach requires a large training set with 3D flame ground truth, which is difficult to obtain. Despite these advancements, both methods still rely on voxel-based flame representation, leading to high computational costs.

Simulation methods are commonly used to analyze flame characteristics by directly modeling the flame through mathematical functions. Chen et al.[4] proposed modeling flames using Gaussian functions. Generally, in this method, the intensity of the flame at a spatial position $x$ can be represented as $g(x)$, which can be seen as a mixture of multiple Gaussian functions.

$$g(X) = \sum_i w_i \frac{1}{\sqrt{(2\pi)^3 |\Sigma_i|}} e^{-\frac{1}{2}(x-\mu_i)^T \Sigma_i^{-1} (x-\mu_i)} \quad (1)$$

Inspired by this, we introduce a novel 3D Gaussian representation to flame. We represent the flame volume using 3D Gaussians, an efficient and optimizable explicit structure proposed by Kerbl et al. The flame is defined as a set of anisotropic Gaussians in the 3D world, and 2D projection images are used as supervision. This method achieves better performance in terms of speed, quality, and storage efficiency.

## 3. METHOD

Flames can be characterized as an assemblage of luminous chemical substances exhibiting heterogeneous shapes and spatial distributions. The process of flame reconstruction is basically equivalent to mapping its spatially varying luminous intensity. To account for flame's non-uniformity, we utilize a multi-Gaussian fitting approach by minimizing the 2D projection errors, as illustrated in Figure 1.

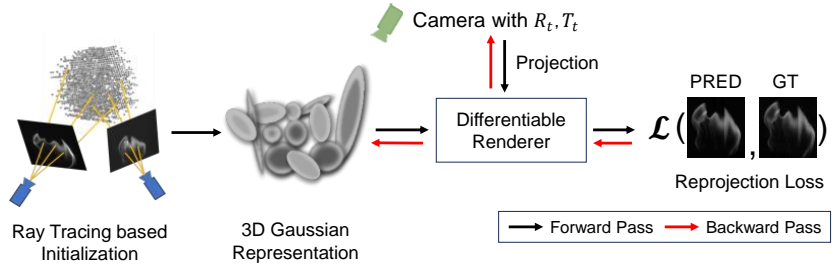

Figure 1. Method overview: We first initialize the flame using a reverse ray tracing method based on several 2D flame projections. Subsequently, we represent the flame using 3D Gaussian functions. By leveraging differentiable volume rendering, we minimize the reprojection error and jointly optimize the 3D Gaussian flame representation with the calibrated camera poses.

### 3.1 3D Gaussian Representation for Flame

We begin by initializing the 3D Gaussian representation using point clouds derived from inverse ray tracing (details in Section 3.2). Each point in the cloud is then represented as an anisotropic Gaussian (akin to an ellipsoid), defined by three key parameters: a position μ, an anisotropic covariance Σ defining its shape, and luminous intensity σ. Consequently, each 3D Gaussian is expressed mathematically as follows [5]:

$$G_i(x) = e^{-\frac{1}{2}(x-\mu_i)^T \Sigma_i^{-1}(x-\mu_i)} \quad (2)$$

Given that flame chemiluminescence directly reflects intensity, we focus exclusively on luminance characteristics to capture the flame's appearance. To account for view-dependent variations, we incorporate spherical harmonics, which model the appearance of only a single channel.

To ensure positive definiteness of the covariance matrix, the covariance $\Sigma$ is decomposed into a 3-dimensional vector $S$ representing the Gaussian's scaling and a quaternion $R$ representing its rotation.

$$\Sigma = RSS^T R^T \quad (3)$$

To render 3D Gaussians to 2D image, the 3D Gaussian's $\mu$ and $\Sigma$ are projected onto the 2D image plane via the splatting process [5]:

$$\mu' = KW\mu \qquad \Sigma' = JW\Sigma W^T J^T \quad (4)$$

Where $K$ is the camera intrinsic parameters, $W$ is the projection matrix used to convert the world coordinate system to the camera coordinate system, and $J$ is the Jacobian of the affine approximation of the projective transformation.

The grayscale value of each pixel in a rendered 2D image (calculated as the integral along a ray passing through the flame) is determined using alpha-blending:

$$C(p) = \sum_{i \in N} c^i \alpha_i \prod_{j=1}^{i-1}(1-\alpha_i), \ \alpha_i = \sigma_i e^{-\frac{1}{2}(\mathbf{p}-\mu_i)^T \Sigma'(\mathbf{p}-\mu_i)} \quad (5)$$

Where $p$ represents the coordinates in the pixel coordinate system, and $c^i$ represents the calculated view-dependent flame grayscale value.

Optimization involves comparing the rendered 2D projection with the actual flame projection image, allowing parameter updates for each 3D Gaussian via differentiable rendering. To further improve reconstruction quality and avoid over-or under-reconstruction, we incorporate adaptive control techniques as prior work[5], such as pruning low-opacity Gaussians and densifying regions during optimization.

### 3.2 Ray Tracing Based Initialization

As previously noted, a 3D Gaussian is defined by a set of optimizable parameters, and their initialization is a critical step before optimization. Existing methods[5] typically initialize these parameters using a point cloud generated through motion recovery techniques.

Structure from Motion (SfM) generally relies on hundreds of images to produce a sparse point cloud. However, for flame reconstruction, capturing such a large number of 2D projection images is impractical. Real-world setups are often limited to around 10 cameras, resulting in significantly sparser data compared to typical 3D reconstruction scenarios. Extremely sparse configurations are beyond the scope of this work, as they lack sufficient information for a good flame reconstruction.

Sparse viewing angles result in minimal overlap, reducing matchable feature points and complicating triangulation for 3D recovery. Moreover, flame projections lack distinct texture features, further hindering feature extraction and matching.

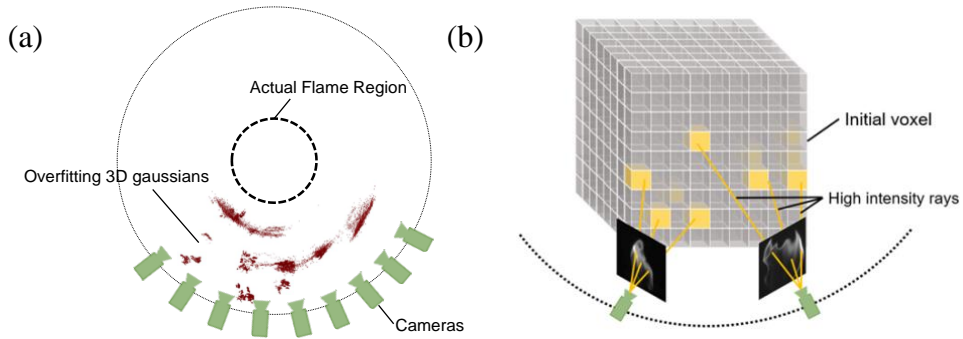

Figure 2. (a) 3D Gaussians (red areas) distribute in front of the cameras to fit the corresponding 2D flame projection images, rather than reconstructing the true 3D flame volume, leading to severe overfitting. (b) Ray-tracing-based random initialization method.

A simple approach, proposed in RAIN-GS[10], is to initialize point clouds randomly over a wide spatial range. However, this method often leads to severe overfitting, with the optimized 3D Gaussians clustering in front of the cameras rather than forming a complete volumetric structure, as shown in Figure 2(a).

To address this limitation, we introduce a novel ray-tracing-based initialization method that constrains the initial distribution of 3D Gaussians. By reversing the flame chemiluminescence process, the cameras are treated as light sources emitting rays. The intersection of these rays from multiple camera views defines the region where the flame volume is likely to exist, as illustrated in Figure 2(b).

Specifically, we generate relatively sparse voxels centered around the camera distribution, with dimensions set to 50×50×50. For image pixels with intensity values exceeding a predefined threshold $\tau$, a ray is emitted from the optical center of the corresponding camera. This ray is then transformed into the world coordinate system using the camera's extrinsic parameters to compute its trajectory.

By tracing rays from multiple viewpoints, we identify intersections that indicate the presence of the flame volume. At these intersection points, 3D Gaussians are initialized. Points without intersecting rays are excluded, ensuring that the initialized point cloud is restricted to the actual flame region. This method effectively confines the initial distribution to areas with sufficient flame information.

### 3.3 Training

*Camera Pose Optimization* Considering the potential for minor calibration errors in the actual camera calibration process, we introduce two optimizable parameters, $\Delta R$ and $\Delta T$, for each training pose. During the optimization process, the camera pose associated with each training instance undergoes the following transformation:

$$\begin{aligned} R'_c &= R_c \Delta R_c \\ T'_c &= T_c + \Delta T_c \end{aligned} \quad (6)$$

Including pose optimization in the rendering pipeline requires calculating transformation gradients during back-propagation. We adopted the gradient derivation approach from previous work[11].

*Loss function* Thus, during the training process, the parameters of each Gaussian and the camera pose variations are jointly optimized using the following loss function:

$$\mathcal{L} = (1 - \lambda_{\text{D-SSIM}})\mathcal{L}_1 + \lambda_{\text{D-SSIM}}\mathcal{L}_{\text{D-SSIM}} \quad (7)$$

Where $\mathcal{L}_1$ refers to $L1$ Loss, $\mathcal{L}_{D-SSIM}$ refers to D-SSIM Loss[5], and the $\lambda_{D-SSIM} = 0.2$.

## 4. EXPERIMENTS

### 4.1 Implementation Details

FlameGS was trained with an RTX 4060 GPU and a 12th Gen Intel Core i7-12650H CPU. Training involved 10,000 optimization iterations. During the first 500 iterations, only photometric loss was optimized. From 500 to 3000 iterations, adaptive density control was applied every 100 iterations, removing Gaussians with opacity below 0.05.

The ART algorithm used for comparison was implemented in Python with a relaxation parameter of 0.01 and 50 iterations.

### 4.2 Experiment Verification

To validate the method, we set up an experimental bench with 10 imaging cameras to capture 2D flame projection images. All camera shutters were synchronized. Each camera captured images at a resolution of $800 \times 1024$. The flame reactor was positioned at the center of the setup. The layout of the devices is illustrated as figure 3 (a).

### 4.3 Quantitative Comparison

We use 10-fold validation in the experiments. We collected a sequence of 100 consecutive frames captured by each camera. For each frame, we performed optimization using 10 images separately. As real flames lack 3D ground truth, comparisons are based on 2D projections. Metrics are calculated using cross-validation, with 9 images for training and 1 for validation per time step, averaged over 10 iterations.

From table 1, our method can reconstruct better results more efficiently. While being approximately 30 times faster in optimization time than the ART algorithm, our method also achieves higher SSIM and PSNR values. Each Gaussian contributes 9 parameters for anisotropic covariance, 3 for position mean, and 1 for luminance. In our experiment, total parameters number in FlameGS stabilizes around a specific number around $1000 \times (9 + 3 + 1) = 13000$. In contrast, the ART method's trainable parameters are fixed to the total number of voxels $120 \times 120 \times 120 = 1,728,000$.

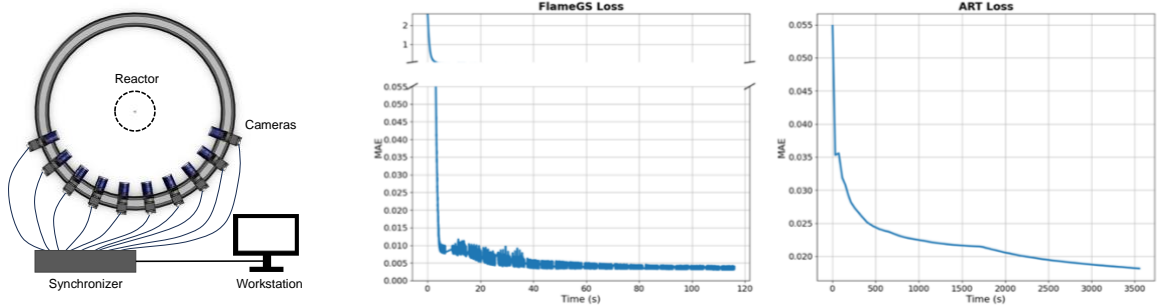

Figure 3. (a) Devices layout. Ten cameras are surrounding the reactor and capture the flame 2D projection simultaneously by using the synchronizer. All data is stored in the workstation. (b) The reconstruction loss with different training steps.

Table 1. Quantitative comparison of different method.

| Model | MAE | PSNR | SSIM | Training time(s) | Memory cost (GB) |
|---|---|---|---|---|---|
| ART | 0.01825 | 38.75 | 0.81 | 3859 | 20.7 |
| FlameGS (Ours) | 0.00453 | 39.05 | 0.96 | 113 | 2.1 |

Figure 3 (b) indicates better training efficiency of our method according to the loss-time plot. At the 120-second mark, FlameGS achieves an MAE close to 0.005, whereas the ART method only reaches this threshold at 3500 seconds. As training time progresses, the MSE Loss of FlameGS decreases rapidly and then reaches a relatively low stable value. Comparing the final MSE Loss of both methods, FlameGS has a lower loss than the ART algorithm.

As shown in Figure 4 (a) and (b), the ART model is constrained by its predefined voxel resolution. Flame volume outside the voxel is ignored by ART. And voxel resolution limits its ability to reconstruct high-frequency details. In contrast, FlameGS accurately reconstructs fine details and aligns with the actual flame's luminous intensity. Additional visual results are provided in Figure 5.

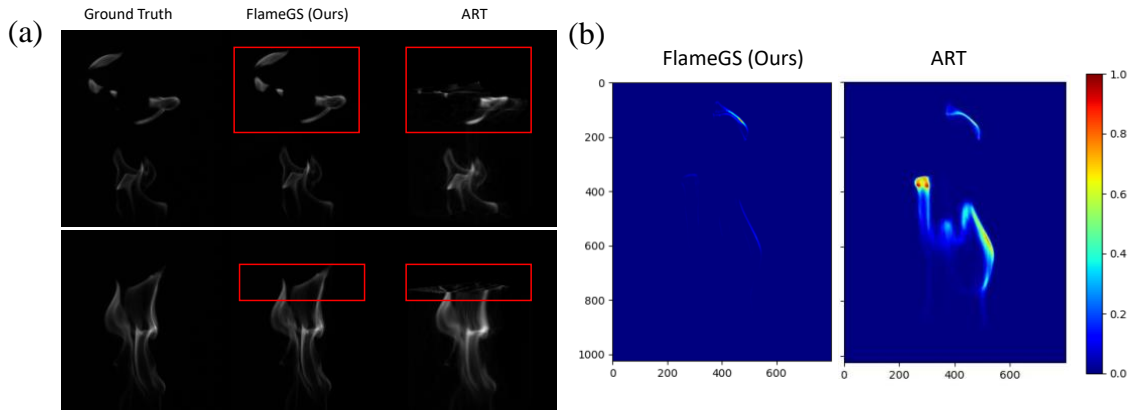

Figure 4. (a) Artifacts caused by predefined voxels in ART method. FlameGS can avoid this issue and can reconstruct the flame in a wider range. (b) Normalized error map of FlameGS and ART against the ground truth image.

# 5. CONCLUSION

We propose a novel flame representation method based on 3D Gaussian, achieving efficient and high-quality flame reconstruction. Additionally, we introduce a ray-trace-based initialization approach that reduces the overfitting problem under sparse viewpoints. Compared to voxel-based reconstruction methods, our approach overcomes the limitation of constrained reconstruction regions and high computation complexity. Our work outperforms the traditional method in both accuracy and efficiency. Future work will explore dynamic flame reconstruction.

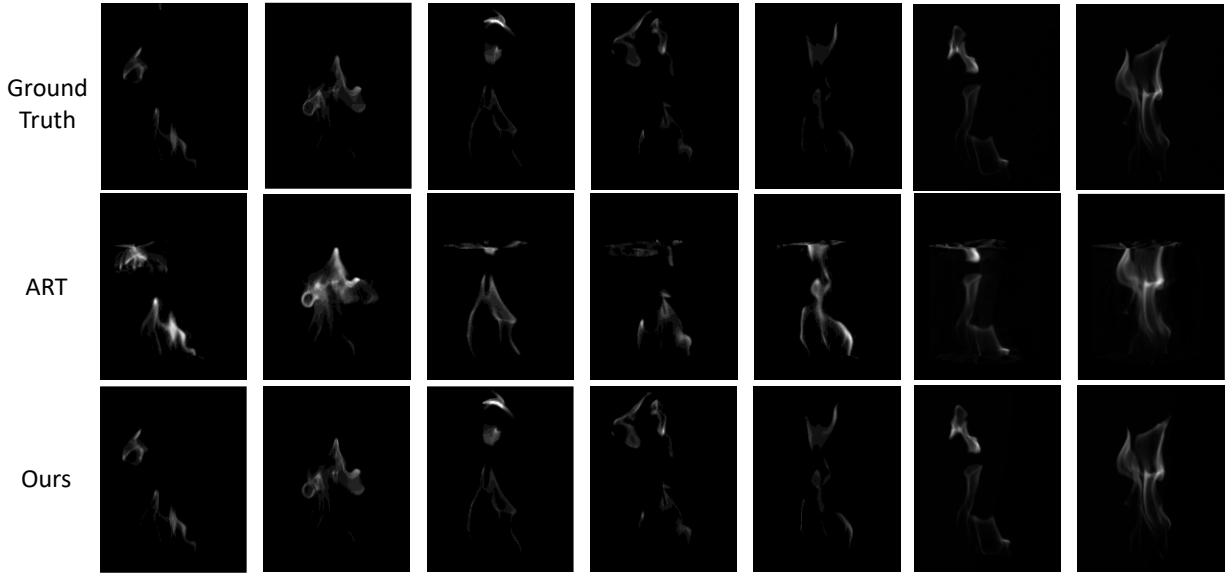

Figure 5. The reconstruction result on the validation view.